# Unsupervised domain adaptation semantic segmentation of high-resolution remote sensing imagery with invariant domain-level prototype memory

Jingru Zhu, Ya Guo, Geng Sun, Libo Yang, Min Deng and Jie Chen, *Member, IEEE*

*Abstract*—**Semantic segmentation is a key technique involved in automatic interpretation of high-resolution remote sensing (HRS) imagery and has drawn much attention in the remote sensing community. Deep convolutional neural networks (DCNNs) have been successfully applied to the HRS imagery semantic segmentation task due to their hierarchical representation ability. However, the heavy dependency on a large number of training data with dense annotation and the sensitiveness to the variation of data distribution severely restrict the potential application of DCNNs for the semantic segmentation of HRS imagery. This study proposes a novel unsupervised domain adaptation semantic segmentation network (MemoryAdaptNet) for the semantic segmentation of HRS imagery. MemoryAdaptNet constructs an output space adversarial learning scheme to bridge the domain distribution discrepancy between source domain and target domain and to narrow the influence of domain shift. Specifically, we embed an invariant feature memory module to store invariant domain-level prototype information because the features obtained from adversarial learning only tend to represent the variant feature of current limited inputs. This module is integrated by a category attention-driven invariant domain-level memory aggregation module to current pseudo invariant feature for further augmenting the representations. An entropy-based pseudo label filtering strategy is used to update the memory module with high-confident pseudo invariant feature of current target images. Extensive experiments under three cross-domain tasks indicate that our proposed MemoryAdaptNet is remarkably superior to the state-of-the-art methods.**

## I. INTRODUCTION

DRIVEN by the rapid growth of Earth observation technology, large amounts of remote sensing imagery with high spatial resolution are increasingly available, which makes Earth observation possible. Automatic interpretation of high-resolution remote sensing (HRS) imagery plays a vital role in the field of remote sensing analysis, such as urban planning, intelligent transportation, agricultural production, and natural disaster monitoring. Specifically, semantic segmentation model is an important tool for the automatic interpretation of HRS imagery. It aims to assign a land cover category (such as building, tree, etc.) to each pixel in the image. In the past 10 years, deep convolutional neural networks (DCNNs) have achieved great success in remote sensing image automatic interpretation tasks, such as scene classification [1, 2], image caption [3, 4], object detection [5, 6], and semantic segmentation [7, 8], due to their excellent capability in exhibiting representations and high-level features. A fully convolutional network (FCN) [9] is a classic semantic segmentation tool based on the DCNN structure. Some studies, including [10, 11], have applied FCNs to remote sensing imagery semantic segmentation task and drawn much attention. More complex encoder–decoder frameworks with skip connection, such as SegNet [12], U-Net [13], PSPNet [14], and Deeplab [15], have been used to HRS imagery semantic segmentation.

However, some major problems are found in applying DCNNs to the semantic segmentation of HRS imagery. (1) The superior performance of DCNN-based semantic segmentation models relies heavily on a massive number of high-quality training samples with dense annotation. Although many HRS semantic segmentation datasets with dense annotation are available in the community, they are often limited to a certain area or an application. Manually annotating the dense semantic label is labor intensive and time consuming. Overall, the insufficient training data limit the availability of DCNN-based semantic segmentation models for HRS imagery. (2) The DCNN-based semantic segmentation methods prefer to be particularly sensitive to the distribution variation. The appearance and structural characteristics of HRS imagery vary because of the diverse imaging conditions, including difference in geolocations, imaging sensors, and observational illumination, as shown in Fig. 1. This condition makes the deep models trained on a certain HRS dataset with annotated labels invalid when dealing with images obtained under different imaging conditions. Therefore, finding an accurate and efficient

Manuscript received ; revised ; accepted ; date of publication ; date of current version. This work was supported by the National Key Research and Development Program of China (Grant 2020YFA0713503) and the National Natural Science Foundation of China (Grant 42071427). (Corresponding author: Jie Chen.)

J. R. Zhu, Y. Guo, G. Sun, L. B. Yang, M. Deng, and J. Chen are with the School of Geosciences and Info-Physics, Central South University, Changsha 410083, China (e-mail: cj2011@ csu.edu.cn).

semantic segmentation method for HRS imagery with the ability to deal with domain gap is crucial.

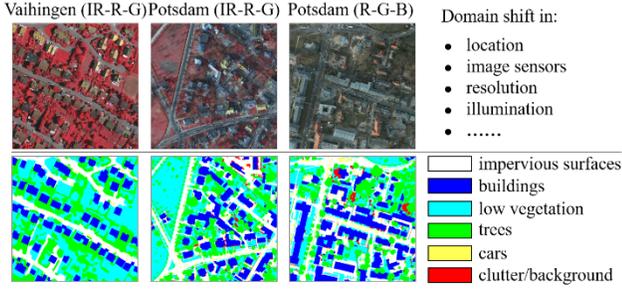

**Fig. 1.** Cross-domain remote sensing imagery.

Unsupervised domain adaptation (UDA) is developed to bridge the domain discrepancy between the source and target domains. UDA can transfer knowledge learned from source domain with dense annotation to unlabeled target domain by reducing the shift of domain distributions. Recently, adversarial learning [16] has raised great attention in the fields of computer vision and has been applied to the UDA semantic segmentation. In accordance with the levels of data that generate adversarial loss, adversarial adaptation methods can be divided into pixel-level, feature-level, and output-level adversarial adaptation methods. The pixel-level adversarial adaptation methods usually implement generative adversarial network-based image translation model to transfer the source domain images into target domain-styled images and then realize the semantic segmentation model on the transferred images [17-19]. They can bridge the domain discrepancy in terms of image style, including illumination, color, and texture. However, they cannot narrow the domain differences in the content structure. Besides, their excellent performance relies on a successful image translation model. The feature-level adversarial adaptation methods focus on the alignment of source and target domains at the feature space through adversarial learning based on the feature representations [20, 21]. However, adversarial learning at high-dimension feature space may be unstable and unreliable due to the heterogeneity and noninterpretability of high-dimension feature. The output-level adaptation methods adopt adversarial learning in the output space [22-24]. The features from the source and target domains can maintain the semantic consistency through a minimax game between segmentation network and discriminator.

Although the above UDA methods based on adversarial learning have achieved impressive performance in semantic segmentation, they are mainly designed for natural images [17, 18, 22, 23]. Given that some differences are found between the remote sensing and natural images in terms of imaging angle, observation subject, scene complexity, and spatial layout, the above methods cannot be directly applied to the cross-domain semantic segmentation of HRS imagery. In recent years, many domain adaptation semantic segmentation methods for HRS imagery have been developed in the remote sensing community [25-28], most which reduce the domain shift by pixel alignment or feature alignment. The existing UDA semantic segmentation methods of HRS imagery need to cut the large-size HRS images into small-size patches, which are inputted into the model in the form of batch ($1 \leq$ batch size $\ll$ dataset size) for the adversarial training. However, the distribution of these batch source or target images cannot represent the data distribution of the entire source or target domain. In other words, the network propagates the gradient based on the adversarial loss generated by these batch images, which causes the extracted features only represent the invariant features of current inputs (batch images of source or target domain) rather than the domain invariant features, which we call pseudo invariant features. Theoretically, a certain bias and variance is observed between pseudo invariant features and domain invariant features, which makes the knowledge learned in the source domain cannot be transferred to the target domain to the greatest extent. We aim that domain adaptation models based on adversarial learning can represent domain invariant features based on all data in the two domains rather than the pseudo invariant features. Therefore, how to obtain domain invariant features from pseudo invariant features is extremely important for cross-domain semantic segmentation of HRS imagery.

Considering these issues, we develop a novel UDA semantic segmentation method for HRS imagery by integrating invariant domain-level prototype information. Given that the output-level adaptation methods can ensure the semantic consistency of features by conducting adversarial learning in the output space, we use an output-level-based adversarial strategy to reduce the distribution difference in the source domain and target domain images, and to obtain the pseudo invariant feature with semantic consistency. As illustrated in Fig. 2, we use a prototype memory module to store the pseudo invariant features of historical source images and target images and dynamically update them to obtain the domain invariant feature in category wise form because a certain bias and variance is found between the pseudo invariant feature and the domain invariant feature. We use an invariant domain-level memory aggregation module to adaptively integrate the domain invariant feature in prototype memory module and the current pseudo invariant feature. Current pseudo invariant feature representations are augmented by integrating the domain invariant feature representations, which enables the model to maximize the transferability of knowledge learned from the source domain to the target domain, thereby enhancing the performance of semantic segmentation model on target HRS imagery. Our main contributions can be summarized as follows:

1) This study develops an invariant domain prototype memory module $M$ to integrate invariant domain-level prototype information for the UDA semantic segmentation of HRS imagery, where the $M$ can be dynamically updated by the pseudo invariant features of historical source images and target images to obtain the invariant domain-level prototype presentation. Concretely, this presentation is stored classwise for the alignment of category-level joint distribution.

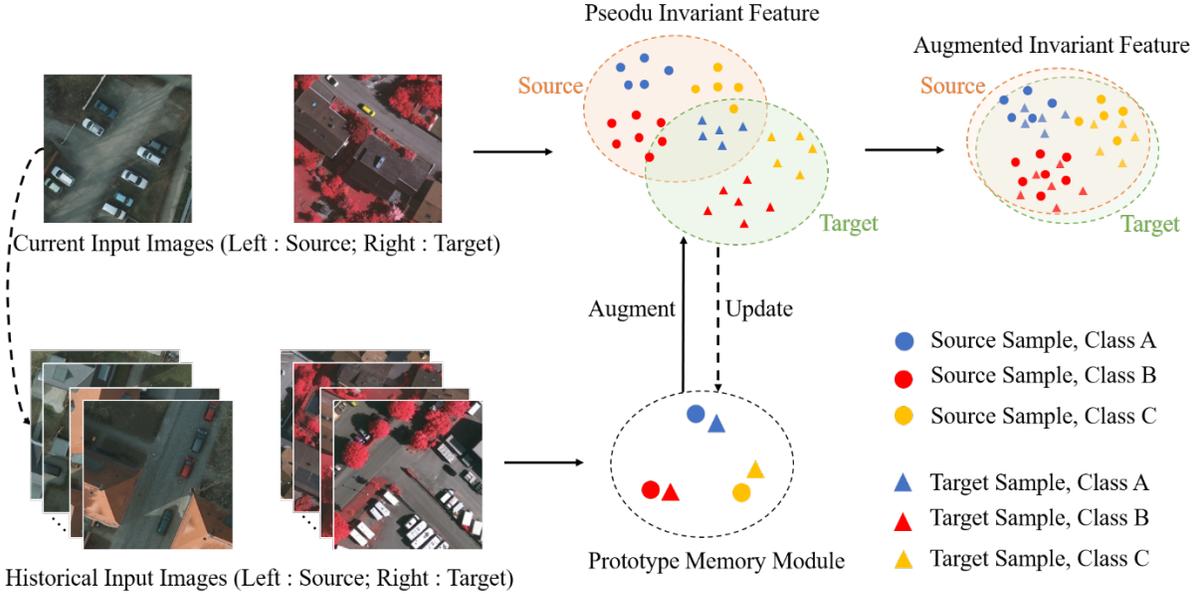

**Fig. 2.** Mining invariant domain-level prototype information beyond current images. The prototype memory module stores the invariant domain-level representations of various categories.

2) This study introduces a category attention-driven invariant domain-level memory aggregation module to better integrate the invariant domain-level prototype information to current pseudo invariant feature for enhancing the feature representations of current input images.

3) This study proposes an entropy-based pseudo label filtering strategy to better utilize the pseudo invariant feature of current target images for dynamically updating the $M$. Specifically, it reserves the prediction in which its entropy is less than or equal to the threshold.

The remainder of this study is organized as follows. Section II summarizes the related work to our study. Section III introduces the proposed method in detail. Section IV discusses the experiment settings, comparative analysis of experimental results, ablation experiments, parameter sensitivity analysis of entropy threshold, and the effects of different data augmentation strategies. Section V provides the conclusion and future research directions.

## II. RELATED WORK

### A. UDA Semantic Segmentation in the Computer Vision Field

UDA can narrow the distribution shift between different data domains, in which we denote the labeled dataset as source domain, and the unlabeled dataset with different distributions from source domain as target domain. Traditional deep adaptation methods [29] used the multiple kernel variant of MMD (MK-MMD) to jointly maximize the difference of source and target domains. As the extension of MMD, [30] explored a CMD to learn the domain-invariant representation by matching the higher order central moments of probability distribution. As another choice of UDA methods, adversarial learning bridges the domain shift by forcing the generator to produce pixels, features or predictions that confuse the discriminator. CyCADA [17] transferred source training data into the target domain by adversarial learning in pixel-level to bridge the low-level discrepancy between the source and target domains. [19] utilized a bidirectional learning framework to promote the image translation model and segmentation adaptation model simultaneously, which can gradually reduce the domain gap. These UDA methods pursue the alignment of source and target domains at the pixel-level. Some adversarial-based UDA methods pursued the adaptation between the source and target domains at the feature space so that the aligned features can generalize to the two domains. [20] used a convolutional domain adversarial training technique to align the distribution of source and target domains at the global-level and category-level. [21] proposed a domain adversarial learning framework joining global and class-specific measures for performing cross-city semantic segmentation task. Considering the heterogeneity and noninterpretability of adversarial-based domain adaptation methods at feature space, the output-level adversarial adaptation methods employed adversarial learning in output space via a minimax game between semantic segmentation network and discriminator. [22] employed output space-based adversarial learning at different feature levels for domain adaptation. [23] used a category-level adversarial learning framework on the output space to enforce the local semantic consistency and global alignment. [24] introduced an adversarial training method based on the entropy of semantic predictions to address the UDA semantic segmentation from the source domain to target domain. The output-level adversarial adaptation methods can ensure the semantic similarity and consistency of source domain and target domain features by adversarial learning in the output space.

Although the adversarial learning-based UDA methods have achieved good performance in the semantic segmentation of natural images, they cannot be directly applied to the UDA semantic segmentation task of HRS imagery because they ignore the special properties of HRS imagery in terms of content complexity, image resolution, and spatial layout.

### B. UDA Semantic Segmentation in Remote Sensing Field

UDA has been proposed to remote sensing for bridging the data shift due to the differences in imaging sensor, geographic location, and atmospheric conditions. Previous UDA semantic segmentation studies of remote sensing mainly focused on scene classification task. [31] considered the denoising autoencoders and domain-adversarial neural networks to learn domain-invariant feature representations and applied them to the UDA scene classification task of hyperspectral and multispectral images. [32] used a subspace alignment method based on DCNN to deal with the UDA task in remote sensing image scene classification. [33] presented a correlation subspace dynamic distribution alignment framework to narrow the distribution difference between the source and target domains for the cross-domain scene classification of remote sensing image. In recent years, increasing researchers have focused on the UDA semantic segmentation task of remote sensing images. [25] designed an appearance adaptation approach with semantic consistency for the UDA semantic segmentation of remote sensing images. To deal with the domain shift issue in road segmentation, [26] presented a stagewise domain adaptation model to align the feature of source and target domains via the interdomain adaptation based on the generative adversarial networks. To learn an excellent semantic segmentation network for UDA remote sensing image semantic segmentation task, [27] exploited an objective function under multiple weakly supervised constraints to minimize the disadvantageous effects of data distribution shift between source and target domains. [28] designed a category-certainty attention to adaptively deal with the unadapted regions and categories for the UDA semantic segmentation of HRS imagery. The input size of the existing UDA methods is usually limited to the batch size ($1 \le$ batch size $\ll$ dataset size) level, which makes the features extracted from the network pseudo invariant features rather than the true domain invariant features, so that the knowledge learned in the source domain cannot be transferred to the target domain to the maximum extent. Therefore, how to obtain the domain invariant features from pseudo invariant features needs to be further studied on the cross-domain HRS imagery semantic segmentation task.

### C. The Memory Module

Memory module can enable the DCNNs to store variables and data beyond a small local range. It has been widely used in various fields of computer vision, such as image captioning [34], video object detection [35], video object segmentation [36], and semantic segmentation [37, 38]. [34] exploited the memory as a context repository of prior knowledge and attached previously generated words to memory for seizing the long-term dependency for personalized image captioning. For video object detection task, MEGA [35] utilized a long range memory module to effectively integrate global and local information, which is vital for recognizing the object in a video. For the semantic segmentation task we focus on in this work, [37] employed a feature memory module to store the dataset-level representations for mining the contextual information more than the current input images. [38] maintained a class wise memory module to yield similar pixel-level feature representations with relevant features from labels data for semi supervised semantic segmentation. In this study, we utilize an effective feature memory module to retain the invariant domain-level representations in category wise form to enable the segmentation network in seizing the domain invariant information more than the current input images and enhance the performance on the target domain image.

## III. METHODOLOGY

We concentrate on the UDA semantic segmentation task of HRS imagery. Given source domain data $X_S$ with $C_K$-class pixel-level labels $Y_S \in \{0, 1, ..., C_K\}^{H \times W}$, where $H$ and $W$ denote the height and width of image data, and $C_K$ represents the number of categories. Our goal is to learn a semantic segmentation model that well performs on unlabeled target domain data $X_T$ by transferring the knowledge learned from the labeled source domain data $X_S$ and $Y_S$.

As shown in Fig. 3, our proposed MemoryAdaptNet makes up a domain feature alignment (DFA) branch and an invariant domain-level memory aggregation (IDMA) branch. It uses the DFA branch to adapt the features between the source and target domain images through adversarial learning in the output space, which is presented in Section III-A. In IDMA, we use a memory module to store each pseudo invariant feature of the current input generated by DFA and dynamically update them to obtain invariant domain-level prototype representation, which is integrated to the current feature representations for further refining the prediction of the final class probability distribution, as introduced in Section III-B. When updating the invariant domain-level prototype presentation in the IDMA branch with pseudo invariant feature of current target images, we use the entropy-based pseudo label filtering strategy to provide a high-confidence pseudo label for the target domain images, which contribute to the calculation of each target category representation for invariant domain-level presentation. The details are provided in Section III-C.

### A. DFA

Domain adaptation algorithms based on the adaptation in output space are one of the effective methods to solve the domain shift between two domains, which can preserve the semantic consistency of source domain and target domain features by performing adversarial learning in the output space. On this basis, we construct an adversarial learning scheme on the output space to align the features between the source and target domains in our proposed framework. Intuitively, the overview of DFA branch is visually shown in Fig. 4.

As shown in Fig. 4, the DFA contains a feature extractor $F$, a classifier $C_1$, and a discriminator $D$, where $F$ and $C_1$ make up segmentation network $G_1$. $F$ extracts features from source and target images, and can be any CNN-based network. To realize a high-quality segmentation performance, we utilize the ResNet-101 [39] pretrained on ImageNet [40] as our backbone for feature extractor $F$. It composes 33 residual blocks, each of which has 3 convolutional layers with a skip connection. We apply an atrous convolution layer with rate = 2 to replace the ordinary convolution layer with kernel $3 \times 3$ in the last residual blocks for extracting dense feature maps and capturing long range context. The detailed structure of the applied ResNet-101 is shown in Fig. 5 (a); $C_1$ classifies the features yielded from $F$ into predefined semantic categories pixel by pixel, such as building, tree, and car. Fig. 5 (b) shows the detailed structure of classifier $C_1$. It consists of an atrous spatial pyramid pooling layer with dilations of 6, 12, and 18 to effectively capture multi-scale information in HRS imagery and two convolution-BatchNorm-ReLU layers to obtain the pixel-level prediction with predefined semantic categories. We utilize an interpolation algorithm to fit the size of pixel-level prediction and input image; $D$ attempts to distinguish whether the input comes from the source domain or target domain, which encourages $F$ to generate feature distributions with semantic consistency between the source domain and the target domain. As show in Fig. 5 (c), $D$ is constructed by five $4 \times 4$ convolution layers with stride of 2, and its channel number is [64, 128, 256, 512, 1].

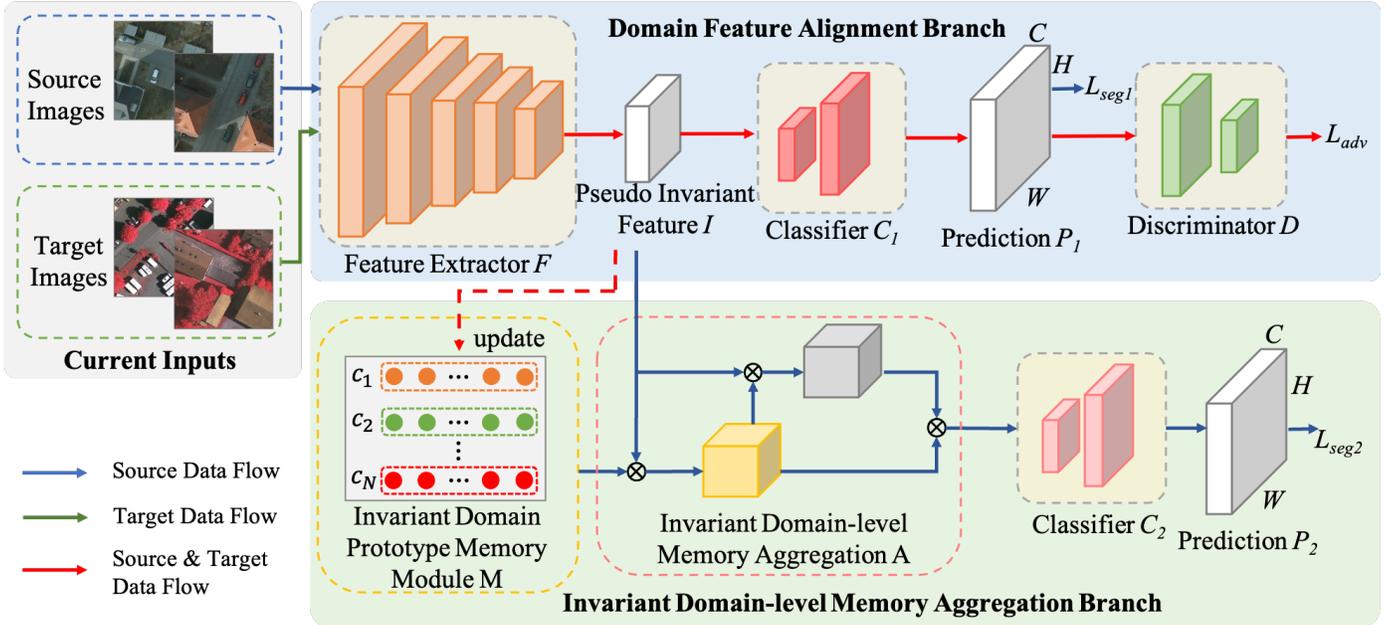

**Fig. 3.** Structure of MemoryAdaptNet.

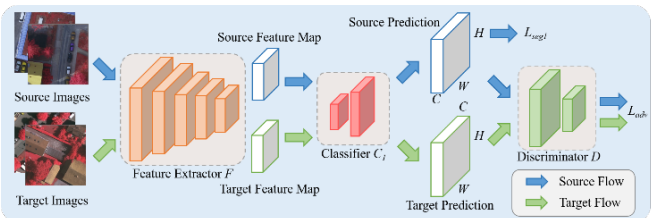

**Fig. 4.** DFA branch.

### B. IDMA

The domain adaptation semantic segmentation method based on adversarial learning normally propagating gradient in each iteration is only based on the adversarial loss generated by the current batch images at each iteration update, causing the extracted features of feature extractor $F$ to represent the invariant features of the current inputs, which cannot represent the invariant features of the two domains. Therefore, the invariant features obtained through the DFA branch belong to pseudo invariant features, preventing the knowledge learned in the source domain to be transferred to the target domain to the greatest extent.

The memory module can enable the DCNNs to depict the valid information beyond the current input image. Contrary to previous memory modules [41, 42] that saved image-level features, classwise memory modules used in [37, 38] have shown their advantage in semantic segmentation task by storing per-pixel features in a class wise form. In this study, the classwise memory module is taken to convert the pseudo invariant feature obtained from the DFA branch to domain invariant feature. Specifically, an IDMA branch is embedded into the DFA branch. As shown in Fig. 3, it makes up an invariant domain prototype memory module $M$, an invariant domain-level memory aggregation module $A$, and a classifier $C_2$, where $F$ $M$, $A$, and $C_2$ make up the segmentation network $G_2$. $M$ is used to store pseudo invariant feature of current input images and update them dynamically in real time. Given that $M$ integrated the pseudo invariant feature of all source and target domain images during dynamic update, the feature stored in $M$ can approximate invariant features of source and target domains. Specifically, $M$ stored the feature in a classwise format to enhance intraclass consistency and interclass difference of HRS imagery. $A$ is designed to integrate the invariant domain-level prototype information in $M$ to current feature for enhancing feature representations of current input images. We use classifier $C_2$ to classify the enhanced features into a precise pixel-level prediction with predefined semantic classes. The details of $M$ and $A$ are presented in Sections III-B-1) and -2).

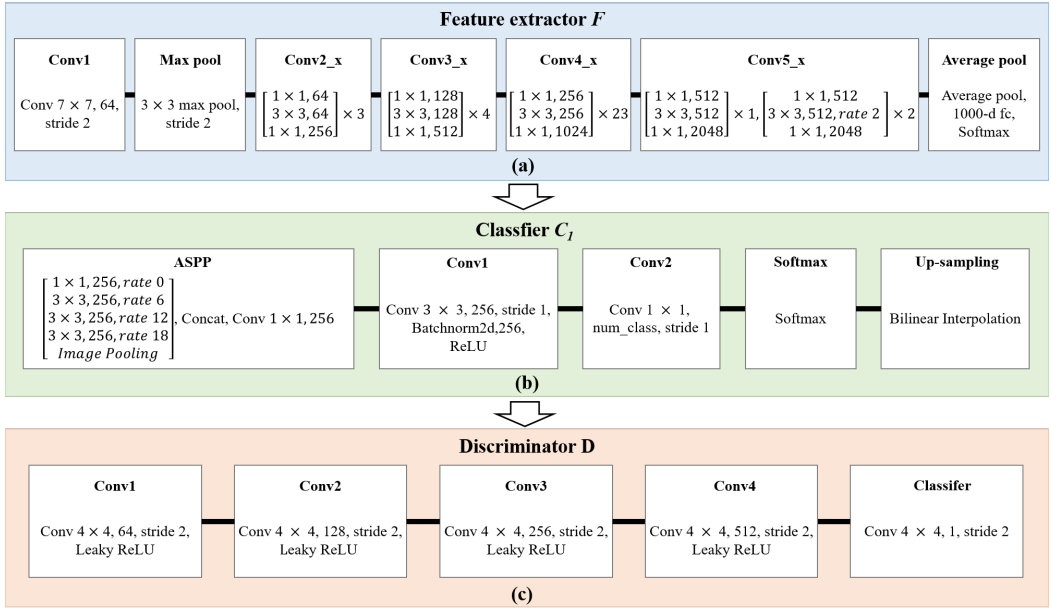

**Fig. 5.** Structure of DFA branch. Different modules are represented by different colors (blue represents feature extractor $F$, green represents classifier $C_1$, orange represents discriminator $D$).

### 1) Invariant Domain Prototype Memory Module

$M$ stores the invariant domain-level representations in category format to enable the network depict invariant-feature information more than the current input images. $M$ is a data vector $\mathcal{M}$ with size $C_K \times C' \times 1$, where $C'$ is the vector size of each category. During training, same as [37, 38], we first calculate the average value for each category from a randomly feature generated by $F$ to initialize $\mathcal{M}$. $\mathcal{M}$ is then updated on every training iteration with a pseudo invariant feature subset of $F_P = (f_p|f_s, f_t \in R^{C \times H \times W})$ generated by $F$. Specifically, given the current pseudo invariant feature $f_p$, we first permute it as size $HW \times C$, that is, $R^{HW \times C}$. Subsequently, we calculate the representation of each category $c_k$ existing in $f_p$,

$$R_{c_k} = \{R^{HW \times C} | GT = c_k\} \in R^{N_{c_k} \times C}, \quad (1)$$

where $R_{c_k}$ is the feature representations of each category $c_k$, $GT \in R^{HW}$ denotes the ground truth with category labels, and $N_{c_k}$ is the number of pixels labeled as $c_k$ in $f_p$. We calculate the similarity $S_{c_k}$ between $R_{c_k}$ and memory of each category $\mathcal{M}_{c_k}$ by cosine similarity.

$$S_{c_k} = \frac{R_{c_k} \cdot \mathcal{M}_{c_k}}{\|R_{c_k}\|_2 \cdot \|\mathcal{M}_{c_k}\|_2}, \quad (2)$$

where $\mathcal{M}_{c_k} \in 1 \times C'$, and $\|\cdot\|_2$ represents the L$_2$-norm. We update the representation of $c_k$ as

$$R'_{c_k} = \sum_{i=1}^{N_{c_k}} \frac{1 - S^i_{c_k}}{\sum_{j=1}^{N_{c_k}} (1 - S^j_{c_k})} \cdot R^i_{c_k}, \quad (4)$$

We calculate the updated memory of each category by leveraging moving average

$$\mathcal{M}'_{c_k} = (1 - m) \cdot \mathcal{M}_{c_k} + m \cdot R'_{c_k}, \quad (5)$$

where $m$ is the momentum, and we employ polynomial annealing policy to update it

$$m_t = (1 - \frac{t}{T})^p \cdot \left(m_0 - \frac{m_0}{100}\right) + \frac{m_0}{100}, t \in [0, T], \quad (6)$$

where $m_t$ represents the value of $m$ at the $t$th iteration, $T$ represent the total number of training iterations, and $p$ and $m_0$ are default parameters and set to 0.9.

### 2) Category Attention-driven Invariant Domain-level Memory Aggregation Module

The $A$ aims to adaptively integrate invariant domain-level prototype stored in $M$ to current feature representations and obtain an enhanced feature. In order to avoid introducing uncertain and incorrect information for the prediction of semantic segmentation during integrating the invariant domain-level category prototypes stored in $M$ into the current pseudo-invariant features of target domain, we design a category attention mechanism to realize the feature aggregation, and Fig. 6 shows the detailed structure. Given the memory feature $\mathcal{M} \in R^{C_K \times C' \times 1}$ stored in $M$ and current pseudo invariant feature $F_P = \{f_p|f_s, f_t\} \in R^{C \times H \times W}$, where C denotes the number of channels in $F_P$. As shown in Fig. 6, three 1×1 convolutional layers are applied to them and generate three features $Q \in R^{C \times H \times W}$, $K \in R^{C_K \times C \times 1}$, and $V \in R^{C_K \times C \times 1}$, respectively. Subsequently, we matrix multiply the reshaped feature $Q$ and feature $K$. A SoftMax function is utilized to compute the affinity attention map $S \in R^{HW \times C_K}$ between each category prototype in memory feature and the current pseudo invariant feature

$$s_{i,j} = \frac{exp(Q_i \cdot K_j)}{\sum_{i=1}^{C_K} exp(Q_i \cdot K_j)}, \quad (7)$$

where $s_{i,j} \in S$ measures the correlation between the $i^{th}$ category memory of feature $K$ and $j^{th}$ channel of feature $Q$, $i = [1,2,\cdots,C_K]$, $j = [1,2,\cdots C]$. We weighted sum the affinity attention map $S$ and feature $V$ to obtain the memory feature $V$'s attention map $S'$ on the current pseudo invariant feature $F_P$ and reshape it to $R^{C \times H \times W}$. The current pseudo invariant feature $F_P$ and the attention map $S'$ are concatenated to acquire the enhanced feature $F_T$ as follows:

$$F_T = \theta(cat(\varphi(S'), F_P)), \quad (8)$$

where $\varphi(\cdot)$ and $\theta(\cdot)$ represent the mapping functions implemented by 1×1 convolution-BatchNorm-ReLU layer. The enhanced feature $F_T$ can adaptively enhance the presentation of current pseudo invariant features by calculating the correlation between the memory feature and original pseudo invariant features. The segmentation network can seize the prototype information more than the current input images, which can enhance the performance of semantic segmentation network on the target domain HRS imagery.

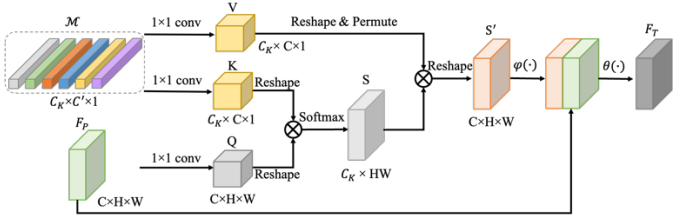

**Fig. 6.** Structure of category attention-driven invariant domain-level memory aggregation module.

## C. Entropy-based Pseudo Label Filtering

In accordance with Section III-B-1, the ground truth is required to calculate the feature representations of each category. However, no ground truth of target domain images is found in the UDA semantic segmentation task of HRS imagery. If the category feature representations of the target domain images are not reserved in the memory module, the memory feature is more inclined to represent the context information of the source domain images and fails to depict the feature distribution of the target domain images, which is detrimental to the semantic segmentation of target domain images. Therefore, the memory module with the pseudo invariant features of the target domain images must be updated. We can use the segmentation predictions of target images as pseudo labels to update the memory module. However, not all pixels in the segmentation predictions are high-confidence, so directly using the segmentation predictions as pseudo labels of target images is unreasonable. On the basis of the observation in Fig. 7, the predictions produced by the model trained only on the source domain have high confidence, low entropy on source images and low confidence, high entropy on target images. Therefore, we propose an entropy-based pseudo label filtering strategy to provide a high-confidence pseudo label for the target domain image. It can assist the updating of target image invariant features in the memory module. This strategy can also make the memory module better represent the invariant domain features, so as to realize superior segmentation performance on the target domain image.

Given the target image set $X_T = \{x_{t,i}\}_{i=1}^{N_t}$, where $N_t$ is the size of target image set. We place each image $x_{t,i}$ to MemoryAdaptNet and obtain a SoftMax prediction set $P_T = \{p_{t,i} \in R^{H \times W \times C_K}\}_{i=1}^{N_t}$ generated by classifier $C_2$, where

$$p_{t,i}^{(h,w)} = \left\{ p_{t,1}^{(h,w)}, p_{t,2}^{(h,w)}, \cdots, p_{t,C_K}^{(h,w)} \right\}, \quad (9)$$

$p_{t,c_k}^{(h,w)}$ represents the probability that pixel $(h, w)$ in $p_{t,i}$ belongs

to class $c_k$, $c_k \in \{1, 2, \cdots, C_K\}$, and $C_K$ is the number of classes.

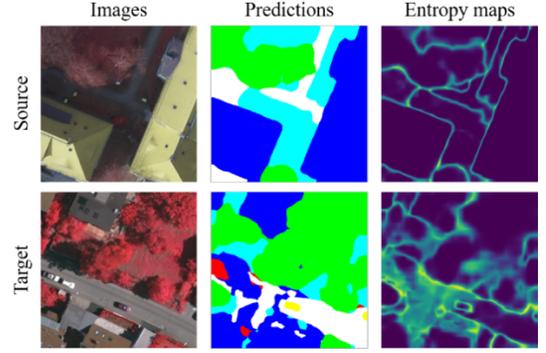

**Fig. 7.** Entropy map of pseudo labels.

We use the probability value to determine whether a pixel is associated with a label, so we calculate the highest probability value of $p_{t,c_k}^{(h,w)}$ by $\mu = max \left( p_{t,c_k}^{(h,w)} \right)$ and assign the index of $\mu$ as the category of pixel $(h, w)$ of $p_{t,i}$. The predicted classification label matrix $l_{t,i}$ can obtained as

$$l_{t,i} \in R^{H \times W}, l_{t,i}^{(h,w)} = index(\mu). \quad (10)$$

Given that the pixels of $l_{t,i}$ are not all high-confidence, we use the entropy map of $l_{t,i}$ to retain the high-confidence pixels and remove the low-confidence pixels. Given the predicted classification probability matrix $p_{t,i}$, the entropy map $E_i \in [0,1]^{H \times W}$ is composed of independent pixelwise entropies

$$E_i^{(h,w)} = \frac{-1}{\log(C_K)} \sum_{c_k=1}^{C_K} p_{t,i}^{(h,w,c_k)} log p_{t,i}^{(h,w,c_k)}, \quad (11)$$

where $E_i^{(h,w)}$ represents the entropy at pixel $(h, w)$ of $E_i$. Specifically, high $E_i^{(h,w)}$ represents low confidence of $l_{t,i}$, whereas low $E_i^{(h,w)}$ represents high confidence of $l_{t,i}$, so we set a threshold $\sigma$ of $E_i^{(h,w)}$ to select the $l_{t,i}$ with high confidence

$$l'_{t,i} = \{R^{H \times W} | E_i \leq \sigma\}. \quad (12)$$

In accordance with Eq. 12, the pseudo label with entropy value greater than the threshold $\sigma$ is discarded, whereas the pseudo label with entropy value less than or equal to the threshold $\sigma$ is reserved.

Owing to the entropy-based pseudo label filtering strategy, we can obtain high-confidence pseudo labels $l'_{t,i}$ of $x_{t,i}$, which makes up the pseudo label set $GT_t = \{l'_{t,i}\}_{i=1}^{N_t}$. $GT_t$ and $GT_s$ make up the ground truth set $GT$, which is used to update the invariant feature memory module.

## D. Network Training

The proposed MemoryAdaptNet model is optimized by two training steps. In step 1, we only train the DFA branch in some epochs for reducing the domain shift between the source and target domains, and obtaining the pseudo invariant feature. Subsequently, we simultaneously optimize the DFA branch and IDMA branch in step 2 to enhance the performance of semantic segmentation network on the target HRS imagery.

### 1) **Step 1:** DFA **Branch Training**

In the first $\tau'$ iterations, we only train the DFA branch. Given a source domain image $x_s \in X_S$ and a target domain image $x_t \in X_T$, we forward them to $F$ and obtain the feature maps $f_s$ and $f_t$, which are inputted to classifiers $C_1$ to acquire the pixel-level predictions $p_s$ and $p_t$, respectively. $p_s$ is used to compute a segmentation loss under the supervision of ground truth $y_s \in Y_S$ for optimizing $G_1$. In this study, we adopt the cross-entropy loss function as the segmentation loss, which is shown as Eq. (13)

$$L_{seg1}(X_S, Y_S) = -E_{(x_s, y_s) \sim (X_S, Y_S)} \sum_{i=1}^{C_K} y_s^{(i)} log G_1(x_s)^{(i)}, \quad (13)$$

where $L_{seg1}(X_S, Y_S)$ is the segmentation loss, $x_s \in X_S$ represents the images from source domain, and $y_s \in Y_S$ represents the ground truth corresponding to $x_s$.

In addition to the segmentation loss, we forward $p_t$ to $D$ to yield an adversarial loss for optimizing $G_1$. The adversarial loss can be defined as follows:

$$L_{adv}(X_T) = -E_{x_t \sim P_T(x)}[log D(G_1(x_t))], \quad (14)$$

where $L_{adv}(X_T)$ represents the adversarial loss generated by target images, and $P_T(x)$ represents the distributions of target domains. The network propagates gradients from $D$ to $G_1$, which encourages $F$ to extract the feature distributions with semantic consistency between the target domain and the source domain.

In summary, the training objective for the segmentation network $G_1$ can be extended from Eq. (13) and (14) as

$$L(X_S, X_T) = L_{seg1}(X_S, Y_S) + \lambda_{adv} L_{adv}(X_T), \quad (15)$$

where $\lambda_{adv}$ is the weight used to balance the adversarial loss $L_{adv}(X_T)$, and we set it to 1.

We then optimize $D$ with the pixel-level prediction $p_s$ and $p_t$. Specifically, we forward $p_s$ and $p_t$ to $D$ to output a single scalar, which denotes the probability that the input came from the source domain rather than the target domain (i.e., label 0 for target training sample and 1 for source training sample). The single scalar is used to compute a cross-entropy loss $L_d$ for optimizing $D$. The loss can be written as

$$L_d(X_S, X_T) = -E_{x \sim P_s(x)}[log D(G_1(X_S))] - E_{x \sim P_t(x)}[log (1 - D(G_1(X_T)))]. \quad (16)$$

2) **Step 2:** DFA **Branch and IDMA Branch Training**

After $\tau'$ iterations, we simultaneously optimize the DFA and IDMA branches. The optimization process of DFA branch is the same as Section III-D-1. For the optimization of IDMA branch, we first use the source feature $f_s$ and the target feature $f_t$ obtained from feature extractor $F$ to update the $M$. We then forward $f_s$ and $\mathcal{M}$ in $M$ to $A$ for integrating the invariant domain-level prototype information in $M$ to the current pseudo invariant feature $f_s$, thereby obtaining the enhanced feature of source images. The enhanced feature is inputted to classifiers $C_2$ to generate the pixel-level prediction $p_s'$, which is used to compute the segmentation loss by cross-entropy loss function as follows:

$$L_{seg2}(X_S, Y_S) = -E_{(x_s, y_s) \sim (X_S, Y_S)} \sum_{i=1}^{N} y_s^{(i)} log C_2\big(A(F(x_s), \mathcal{M})\big)^{(i)}, \quad (17)$$

In summary, the total semantic segmentation loss of MemoryAdaptNet can be expressed as

$$L_{seg}(X_S, Y_S) = L_{seg1}(X_S, Y_S) + L_{seg2}(X_S, Y_S). \quad (18)$$

The whole optimization process of MemoryAdaptNet is summarized in Algorithm I for better understanding.

---

**Algorithm I** Optimization process of MemoryAdaptNet

**Input:** Source image $x_s \in X_S$ and target image $x_t \in X_T$;
**Output:** The predicted label $p_s'$ of source image $x_s$ and the predicted label $p_t'$ of target image $x_t$;

1: **for** $\tau = 0$; $\tau < \tau'$; $\tau$++ **do**
2:   Use source image $x_s$ and target image $x_t$ to optimize the DFA branch with Eqs. 15 and 16;
3: **end for**
4: **for** $\tau = \tau'$; $\tau < \mathcal{T}$; $\tau$++ **do**
5:   Simultaneously optimize the DFA branch and IDMA branch;
6:   Using the source feature $f_s$ and the target feature $f_t$ from feature extractor $F$ to update the memory value in invariant feature memory module $M$;
7:   Use source image $x_s$ and target image $x_t$ to optimize the DFA branch with Eqs. 15 and 16;
8:   Use the source feature $f_s$ from feature extractor $F$ to optimize the IDMA branch with Eq.1 7;
9: **end for**

---

## IV. EXPERIMENTS AND RESULT ANALYSIS

In this section, we first present the experimental settings in Section IV-A, including the description of datasets, task settings, implementation details, and evaluation metrics. We conduct comparison experiments with the state-of-the-art UDA semantic segmentation methods that extract the pseudo invariant feature to verify the effectiveness of our proposed network in Section IV-B. In Section IV-C, we perform an ablation study to demonstrate the effectiveness of each module in our MemoryAdaptNet. We conduct parameter sensitivity analysis to select the appropriate $\sigma$ value for entropy-based pseudo label filtering strategy in Section IV-D. We discuss the effects of different data augmentation strategies on the UDA semantic segmentation of HRS imagery in Section IV-E.

### A. Experimental Settings

1) **Datasets**

To demonstrate the importance of MemoryAdaptNet on the UDA semantic segmentation of HRS imagery, we perform our experiments on two very high-resolution datasets: Vaihingen 2D dataset and Potsdam 2D dataset supplied by the International Society for Photogrammetry and Remote Sensing (ISPRS) WG II/4 [43]. All images in the two datasets are provided with semantic labels, which consist of six general land cover categories: Imp. surf. (impervious surfaces), building (buildings), low veg. (low vegetation), tree (trees), car (cars), and clutter (clutter/background). In the training process, we do not employ the semantic labels of the target domain images.

**Vaihingen dataset:** The Vaihingen dataset draws a small village with sparse layout pattern. It contains 33 high-resolution true orthophoto tiles (TOPs) with an average size of 2,494 × 2,064 pixels and a spatial resolution of 9 cm. As shown in Fig. 8 (a), the near-infrared (IR), red (R), and green (G) channels are provided in this dataset. We employ ID: 2, 5, 7, 8, 13, 20, 22, 24 for testing, and the remaining 25 images for training and validation.

**Potsdam dataset:** The Potsdam dataset depicts a city scene with crowded residential pattern. It contains 38 TOPs with a fixed size of 6,000 × 6,000 and a spatial resolution of 5 cm. The dataset provides near-IR, red (R), green (G), and blue (B) channels, which are combined into three channel compositions: [IR, R, G], [R, G, B], and [IR, R, G, B]. We employ ID: 2_13, 2_14, 3_13, 3_14, 4_13, 4_14, 4_15, 5_13, 5_14, 5_15, 6_13, 6_14, 6_15, 7_13 for testing, and the remaining 24 images for training and validation. As shown in Figs. 8 (b) and (c), we employ the [IR, R, G] and [R, G, B] channel compositions in our experiments.

Fig. 9 represents the pixel percentage of each category to the total number of pixels in the Potsdam and Vaihingen training datasets. Concretely, the pixel proportions of imp. surf., building, low veg., tree, car, and clutter in the Potsdam training datasets are 28.46%, 26.72%, 23.54%, 14.62%, 1.69%, and 4.96%, respectively. The pixel proportions of imp. surf., building, low veg., tree, car, and clutter in the Vaihingen training datasets are 28.69%, 26.70%, 20.56%, 21.91%, 1.29%, and 0.86%, respectively. The two datasets have the characteristics of unbalanced category samples. For example, the proportion of car and cluster categories in the two datasets is remarkably lower than other categories.

### 2) Task Settings

We set up three UDA semantic segmentation tasks in our experiments: (1) P2V_S task, where the Potsdam dataset with [IR, R, G] channel composition serves as the source domain dataset, and the Vaihingen dataset with [IR, R, G] channel composition serves as the target domain dataset. (2) P2V_D task, where the Potsdam with [R, G, B] channel composition serves as the source domain dataset, and the Vaihingen dataset with [IR, R, G] channel composition serves as the target domain dataset. (3) V2P task, where the Vaihingen dataset with [IR, R, G] channel composition serves as the source domain dataset, and the Potsdam dataset with [IR, R, G] channel composition serves as the target domain dataset. In the training phase, we crop the Potsdam ([IR, R, G] and [R, G, B] channel compositions) datasets and their paired labels into 512 × 512 size patches with horizontal and vertical strides of 256 pixels and yield a total of 13,310 patches. With regard to the Vaihingen_[IR, R, G] dataset, we crop the images and their paired labels into 512 × 512 size patches with horizontal and vertical strides of 256 pixels and yield approximately 1,700 patches. In P2V_S and P2V_D tasks, we utilize the Potsdam training set for training and the Vaihingen testing and validation sets for testing and validation. In V2P task, we utilize the Vaihingen training set for training and the Potsdam testing and validation sets for testing and validation.

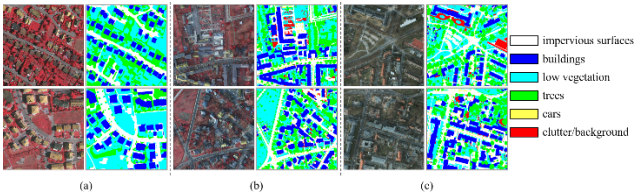

**Fig. 8.** Examples of image and classwise ground truth labels for the ISPRS Vaihingen and Potsdam datasets. (a) is the Vaihingen dataset with [IR, R, G] channel composition, (b) is the Potsdam dataset with [IR, R, G] channel composition, and (c) is the Potsdam dataset with [R, G, B] channel composition.

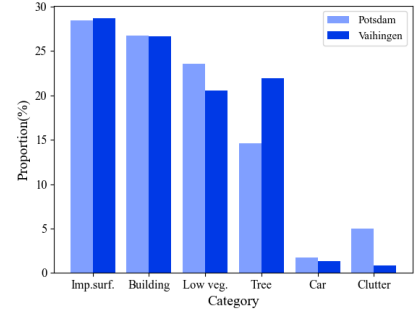

**Fig. 9.** Pixel percentage of each category in Vaihingen and Potsdam training datasets.

### 3) Implementation Details

We implement our network using the PyTorch toolbox on a single RTX 3090 GPU with 24 GB memory. To train the semantic segmentation networks $G_1$ and $G_2$, we employ the stochastic gradient descent optimizer [44] with the initial learning rate of 2.5×10−4, momentum of 0.9, and weight decay of 10−4. For the discriminator training, we employ the Adam optimizer [45] with the initial learning rate of 10−4 and momentum of 0.9 and 0.99.

### 4) Evaluation Metric

We employ four general evaluation metrics, namely, $F_1$-score, OA, MA, and mIoU [9], to assess the performance of different UDA semantic segmentation methods. The higher the values of these metrics, the better the semantic segmentation performance.

$F_1$-score is a common evaluation metric applied to the semantic segmentation task. It can be expressed as follows:

$$F_1 = 2 \times \frac{Precision + Recall}{Precision + Recall}, \quad (19)$$

where the Precision and Recall of category $c_i$ can be computed as follows:

$$\text{Precision}_{c_i} = n_{c_i c_i} / \sum_{c_j} n_{c_j c_i}, \quad (20)$$

$$\text{Recall}_{c_i} = n_{c_i c_i} / \sum_{c_j} n_{c_i c_j}, \quad (21)$$

where $n_{c_i c_i}$ denotes the number of pixels of category $c_i$ predicted as category $c_i$, $n_{c_i c_j}$ denotes the number of pixels of category $c_i$ predicted as category $c_j$, and $n_{c_j c_i}$ denotes the number of pixels of category $c_j$ predicted as category $c_i$.

OA represents the overall accuracy of semantic segmentation performance, MA represents the mean accuracy of semantic segmentation performance, and mIoU represents the intersection over union on each category of predicted labels. They can be expressed as follows:

$$\text{OA} = \sum_{c_i} n_{c_i c_j} / \sum_{c_i} \sum_{c_j} n_{c_i c_j}, \quad (22)$$

$$\text{MA} = (1 / C_K) \sum_{c_i} \left( n_{c_i c_i} / \sum_{c_j} n_{c_i c_j} \right), \quad (23)$$

$$\text{mIoU} = (1/C_K) \sum_{c_i} n_{c_i c_i} / \left( \sum_{c_j} n_{c_i c_j} + \sum_{c_j} n_{c_j c_i} - n_{c_i c_i} \right). \quad (24)$$

### B. Comparative Experimental Result Analysis

To confirm the effectiveness of MemoryAdaptNet with domain invariant features, we perform comparison experiments with a deep model without DA (Source-only) and nine state-of-the-art UDA methods without domain invariant features that performs gradient propagation and parameters update only based on the loss generated by the current batch images (MCD_DA, CLAN, AdaptSegNet, AdvEnt, FADA, SDCA, MUCSS, Zhang's and ProCA), in which MUCSS, Zhang's are the state-of-the-art UDA methods proposed in the remote sensing semantic segmentation, and we directly used the quantitative evaluation results reported in the original paper.

**Source-only:** segmentation network only trained on the source domain and is directly tested on the target domain.

**MCD_DA [46]:** a UDA method that attempts to align source and target domain distributions by considering task-specific decision boundaries.

**CLAN [23]:** a UDA method that aims to enforce the alignment of class-level consistency and global distribution for UDA in semantic segmentation.

**AdaptSegNet [22]:** a UDA method employing multilevel adversarial learning in the output space for pixel-level semantic segmentation.

**AdvEnt [24]:** a UDA method for semantic segmentation via adversarial training approach based on the entropy of semantic predictions.

**FADA [47]:** a UDA method for class-level feature alignment by fine-grained adversarial learning strategy.

**SDCA [48]:** a UDA method that aligns the pixel-wise representation guided by semantic distribution.

**MUCSS [27]:** a UDA method under multiple weakly-supervised constraints to minimize the data shift of remote sensing images.

**Zhang's [49]:** a UDA method exploiting curriculum-style and local-to-global adaptation to address the cross-domain problem of remote sensing images.

**ProCA [50]:** a UDA method that aligns the class-centered distribution by prototypical contrast learning.

For the sake of fairness, all the comparative methods use the DeepLab-v3 with ResNet-101 pretrained on ImageNet as the semantic segmentation network. Three cross-domain experimental tasks are set: P2V_S task, P2V_D task, and V2P task. The batch size and iteration are all set to 2 and 150k, respectively.

#### 1) Comparative Studies on P2V_S Task

To prove the effectiveness of MemoryAdaptNet, we conduct the comparative experiment on the P2V_S task. The domain shift in this task is mainly found in geographic location, spatial resolution, and illumination. The quantitative evaluation results of the compared domain adaptation methods on this domain shift are presented in Table I.

From Table I, due to the effects of domain shift, the source-only model achieves the worst performance with OA, MA, and mIoU values of 67.19%, 60.55%, and 43.58%, respectively.

Compared with the source-only model, the performance of seven comparative state-of-the-art UDA methods achieves improvement in varying degrees by employing different domain adaptation strategy. However, since they only calculate the loss based on the information of current batch images during network parameters update, their features extracted by the feature extractor tend to represent the invariant features of current batch images rather than the domain invariant feature. Our MemoryAdaptNet can convert pseudo invariant features into domain invariant features by considering the information of all domain data and forming category prototype, and achieves the highest OA, MA, and mIoU values with 77.87%, 77.05%, and 56.05%, respectively. Compared with the comparison method MCD_CA with the worst performance, our MemoryAdaptNet shows the performance improvement of 8.01% on OA, 25.35% on MA, and 14.54% on mIoU. Compared with the comparison method ProCA with the best performance, our MemoryAdaptNet presents the improvement of 0.39%, 7.8%, and 1.68% on OA, MA, and mIoU. These performance differences demonstrate that our MemoryAdaptNet can better deal with the domain discrepancy on P2V_S semantic segmentation task by considering the invariant domain-level prototype memory.

In Table I, the $F_1$ scores of MemoryAdaptNet in categories, including low veg., tree, and clutter, realize the best accuracies of 68.22%, 78.52%, and 50.30%, respectively, compared with those of other models, which bring the improvement of 4.54%~15.29% on low veg., 0.03%~8.04% on tree, and 12.54%~40.41% on clutter, respectively. Although our MemoryAdaptNet did not obtain the highest $F_1$ score in the category of imp. surf., building, and car, it still achieved 6.8%, 12.05% and 14.33% improvement compared with source only model. This result shows that our method can relieve the effect of sample category imbalance on the UDA semantic segmentation of HRS imagery by using the effective and representative prototype features of each category stored in the memory module.

Fig. 10 shows the qualitative results on the P2V_S task, where the segmentation results of source-only model show great difference with ground truth due to the serious domain shift problem. The segmentation performance is improved to varying degrees after adaptation. However, the semantic segmentation performance of the four state-of-the-art methods on low veg., car, and clutter categories is poor. For example, a semantic confusion is observed between clutter, car, low veg., and imp. surf. The segmentation boundary of ground objects is rough. Compared with other domain adaptation methods, our MemoryAdaptNet achieves the best semantic segmentation performance, in which the segmentation performance for each category and the problems of semantic errors and unsmoothed boundary are improved.

#### 1) Comparative Studies on P2V_D Task

In this section, we implement the comparative experiment on the P2V_D task. In addition to the difference in geographical location, spatial resolution, and illumination, the domain shift in this task includes the difference in imaging sensors, which has a larger domain gap than the P2V_S task. The quantitative evaluation results of the compared domain adaptation methods on this domain shift are presented in Table II.

TABLE I
QUANTITATIVE EVALUATION RESULTS (%) OF DIFFERENT UDA MODELS ON THE P2V_S TASK

| Methods | | $F_1$ score | | | | | OA | MA | mIoU |
| | Imp. surf. | Building | Low veg. | Tree | Car | Clutter | | | |
|---|---|---|---|---|---|---|---|---|---|
| Source-only | 72.85 | 74.40 | 56.42 | 67.96 | 46.76 | 28.49 | 67.19 | 60.55 | 43.58 |
| State-of-the-art methods | | | | | | | | | |
| MCD_DA (2018) | 77.00 | 80.30 | 54.94 | 70.48 | 54.28 | nan | 70.20 | 57.66 | 46.63 |
| CLAN (2019) | 75.68 | 83.88 | 60.76 | 74.40 | 41.28 | 9.89 | 71.62 | 58.37 | 44.02 |
| AdaptSegNet (2018) | 77.69 | 85.99 | 60.97 | 74.06 | 52.01 | 26.43 | 73.70 | 64.00 | 48.65 |
| AdvEnt (2019) | 77.55 | 84.42 | 52.93 | 74.23 | 53.98 | 30.51 | 74.43 | 65.29 | 51.37 |
| FADA (2020) | 78.77 | 87.84 | 60.84 | 78.31 | 57.82 | 37.76 | 76.60 | 67.72 | 52.55 |
| SDCA (2021) | 81.80 | 88.55 | 56.47 | 77.29 | **64.19** | 31.20 | 76.71 | 67.77 | 52.79 |
| MUCSS (2021) | 66.13 | 76.77 | 55.97 | 73.14 | 51.09 | 45.65 | 61.43 | - | 45.38 |
| Zhang's (2022) | 80.13 | 86.65 | 64.16 | 71.90 | 61.94 | 31.34 | 66.02 | - | 52.03 |
| ProCA (2022) | **82.42** | **90.31** | 63.68 | 78.49 | 56.54 | 37.45 | 77.48 | 69.25 | 54.37 |
| MemoryAdaptSegNet (ours) | 79.65 | 86.45 | **68.22** | **78.52** | 61.09 | **50.30** | **77.87** | **77.05** | **56.05** |

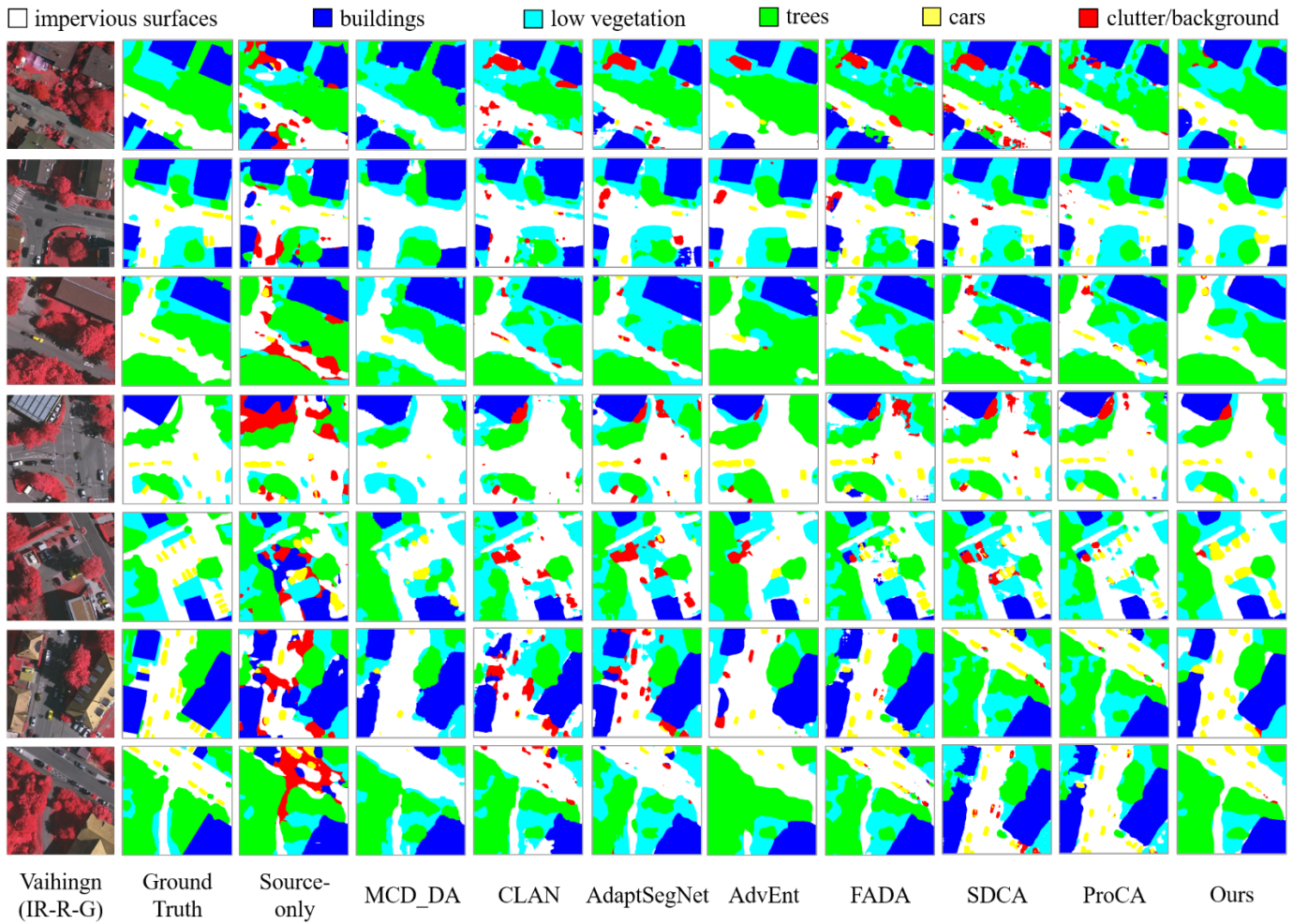

**Fig. 10.** Qualitative results on the P2V_S task.

From Table II, the source-only model achieves the worst performance with OA, MA, and mIoU values of 57.38%, 48.07%, and 33.58%, respectively. The performance of semantic segmentation is improved after some domain adaptations. Although our method does not achieve the highest

OA values, it reaches the highest MA and mIoU values of 71.51% and 46.88%, respectively, which obtains the improvement of 23.44% and 13.30% on MA and mIoU compared with the source-only model. Our proposed method achieves the highest $F_1$ score in the categories of low veg., car, and clutter with the values of 57.26%, 58.35%, and 35.97%, which provides an improvement of 13.72%, 17.91%, and 28.03%, respectively, compared with the source-only model. These results show that

TABLE II
QUANTITATIVE EVALUATION RESULTS (%) OF DIFFERENT UDA MODELS ON THE P2V_D TASK

| Methods | | $F_1$ score | | | | | OA | MA | mIoU |
|---|---|---|---|---|---|---|---|---|---|
| | Imp. surf. | Building | Low veg. | Tree | Car | Clutter | | | |
| Source-only | 64.91 | 63.58 | 43.54 | 63.97 | 40.44 | 7.94 | 57.38 | 48.07 | 33.58 |
| State-of-the-art methods | | | | | | | | | |
| MCD_DA (2018) | 69.75 | 75.46 | 51.62 | 50.85 | 35.05 | 0.00 | 61.75 | 48.30 | 36.52 |
| CLAN (2019) | 67.67 | 77.04 | 47.26 | 58.96 | 26.76 | 8.43 | 62.94 | 50.35 | 35.96 |
| AdaptSegNet (2018) | 68.31 | 78.23 | 46.27 | 63.73 | 40.44 | 8.09 | 63.13 | 51.08 | 37.55 |
| AdvEnt (2019) | 69.49 | 75.87 | 32.96 | 73.72 | 49.79 | 6.39 | 62.04 | 52.80 | 37.64 |
| FADA (2020) | **79.84** | 85.32 | 25.44 | **76.14** | 56.91 | 14.52 | 69.37 | 66.32 | 44.08 |
| SDCA (2021) | 70.16 | 87.05 | 41.42 | 74.89 | 47.79 | 30.25 | 69.48 | 59.99 | 44.38 |
| MUCSS (2021) | 61.33 | 83.00 | 42.17 | 70.66 | 57.88 | 13.88 | 54.82 | - | 39.93 |
| Zhang's (2022) | 76.89 | 84.81 | 56.26 | 68.10 | 57.15 | 18.64 | 60.31 | - | 46.13 |
| ProCA (2022) | 76.77 | **88.71** | 42.93 | 64.38 | 50.97 | 20.67 | **71.39** | 64.07 | 45.63 |
| MemoryAdaptSegNet (ours) | 72.59 | 81.01 | **57.26** | 69.29 | **58.35** | **35.97** | 69.60 | **71.51** | **46.88** |

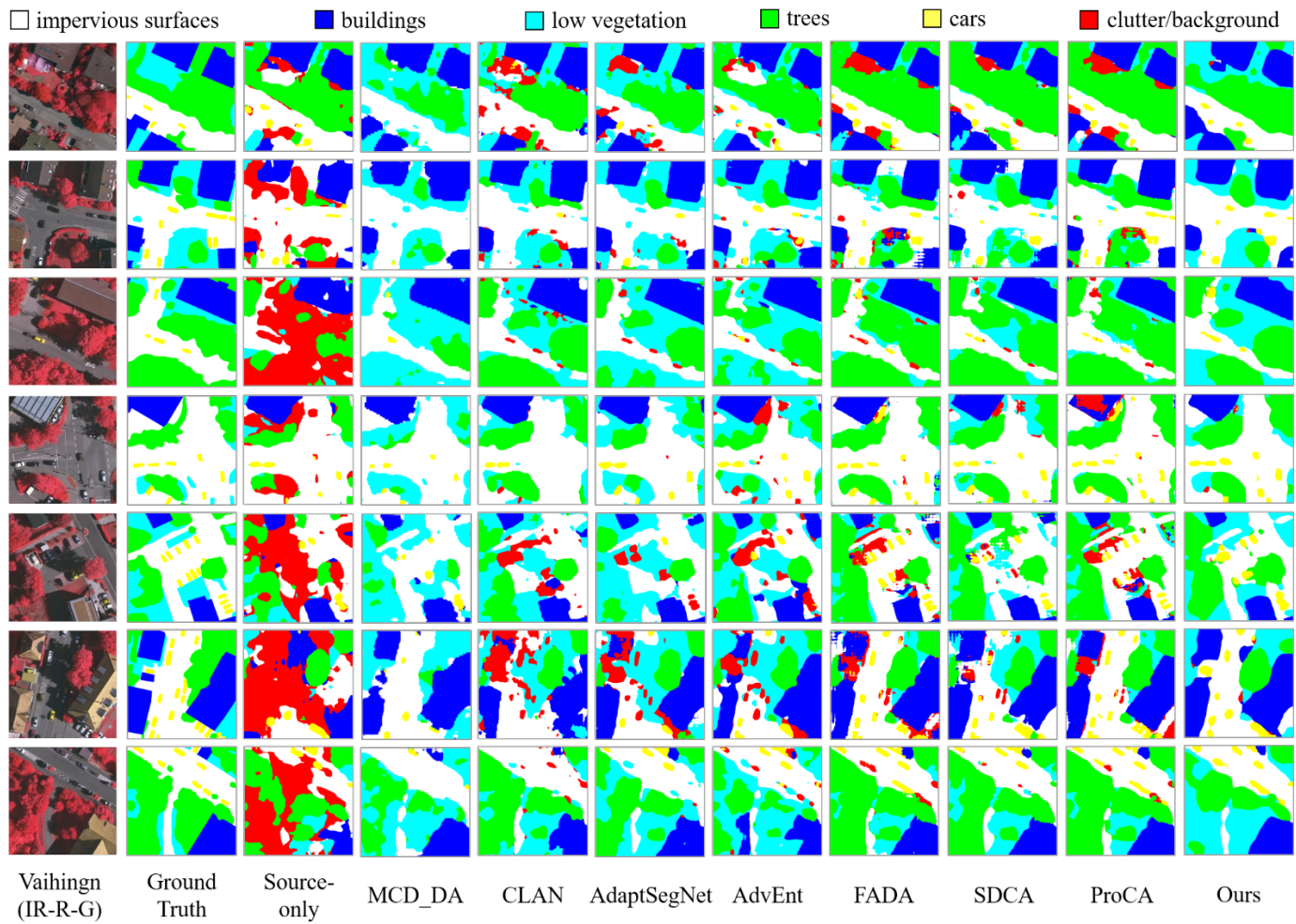

■ impervious surfaces ■ buildings ■ low vegetation ■ trees ■ cars ■ clutter/background

Vaihingn (IR-R-G) | Ground Truth | Source-only | MCD_DA | CLAN | AdaptSegNet | AdvEnt | FADA | SDCA | ProCA | Ours

**Fig. 11.** Qualitative results on the P2V_D task.

our MemoryAdaptNet can deal well with the domain shift in terms of geographic location, spatial resolution, and imaging sensor.

Fig. 11 represents the qualitative results of compared domain adaptation methods on the P2V_D task, where the source-only model achieves the worst semantic segmentation performance. Although the semantic segmentation performance of the four state-of-the-art methods is improved after domain adaptation, it still has some shortcomings in the aspects of unsmooth boundary and semantic confusion. The offset of the input limited data prevents the extracted features to represent the true domain invariant feature, which results in the knowledge learned in the source domain cannot be transferred to the target domain to the greatest extent. Our MemoryAdaptNet represents the domain variant features by updating the memory module with the historical pseudo invariant features of source domain and target domain images, enabling the classifier to classify by learning the invariant domain-level presentation and the feature presentation of the current input image, thereby enhancing the performance of semantic segmentation network on the target HRS images. Fig. 11 confirms the effectiveness of our model on the P2V_D task.

TABLE III
QUANTITATIVE EVALUATION RESULTS (%) OF DIFFERENT UDA MODELS ON THE V2P TASK

| Methods | | $F_1$ score | | | | | OA | MA | mIoU |
|---|---|---|---|---|---|---|---|---|---|
| | Imp. surf. | Building | Low veg. | Tree | Car | Clutter | | | |
| Source-only | 71.32 | 69.45 | 60.97 | 54.67 | 69.10 | 5.47 | 64.36 | 54.05 | 41.22 |
| State-of-the-art methods | | | | | | | | | |
| MCD_DA (2018) | 75.67 | 76.72 | 63.53 | 39.97 | 60.68 | 0.00 | 67.04 | 56.42 | 43.71 |
| CLAN (2019) | 75.94 | 79.08 | 64.44 | 49.17 | 66.91 | 6.77 | 67.87 | 57.10 | 44.12 |
| AdaptSegNet (2018) | 76.55 | 77.93 | 62.15 | 45.45 | 70.89 | 6.16 | 68.93 | 57.33 | 44.94 |
| AdvEnt (2019) | 76.65 | 82.64 | 63.73 | 32.94 | 71.74 | 2.76 | 69.59 | 58.03 | 45.43 |
| FADA (2020) | 79.04 | **82.89** | 69.31 | 29.02 | **82.75** | 5.82 | 71.20 | 59.63 | 46.75 |
| SDCA (2021) | 79.35 | 82.17 | 70.10 | 41.35 | 78.76 | 5.06 | 71.74 | 60.03 | 47.18 |
| MUCSS (2021) | 67.53 | 69.59 | 53.48 | 51.82 | 65.31 | 20.56 | 54.71 | - | 39.30 |
| Zhang's (2022) | 78.59 | 79.84 | 63.27 | 54.60 | 75.08 | **24.59** | 62.66 | - | 47.87 |
| ProCA (2022) | 77.81 | 81.20 | **71.94** | 37.90 | 81.78 | 15.56 | 71.90 | 60.52 | 48.20 |
| MemoryAdaptSegNet (ours) | **79.70** | 77.84 | 67.58 | **64.06** | 76.54 | 16.15 | **72.25** | **61.87** | **49.82** |

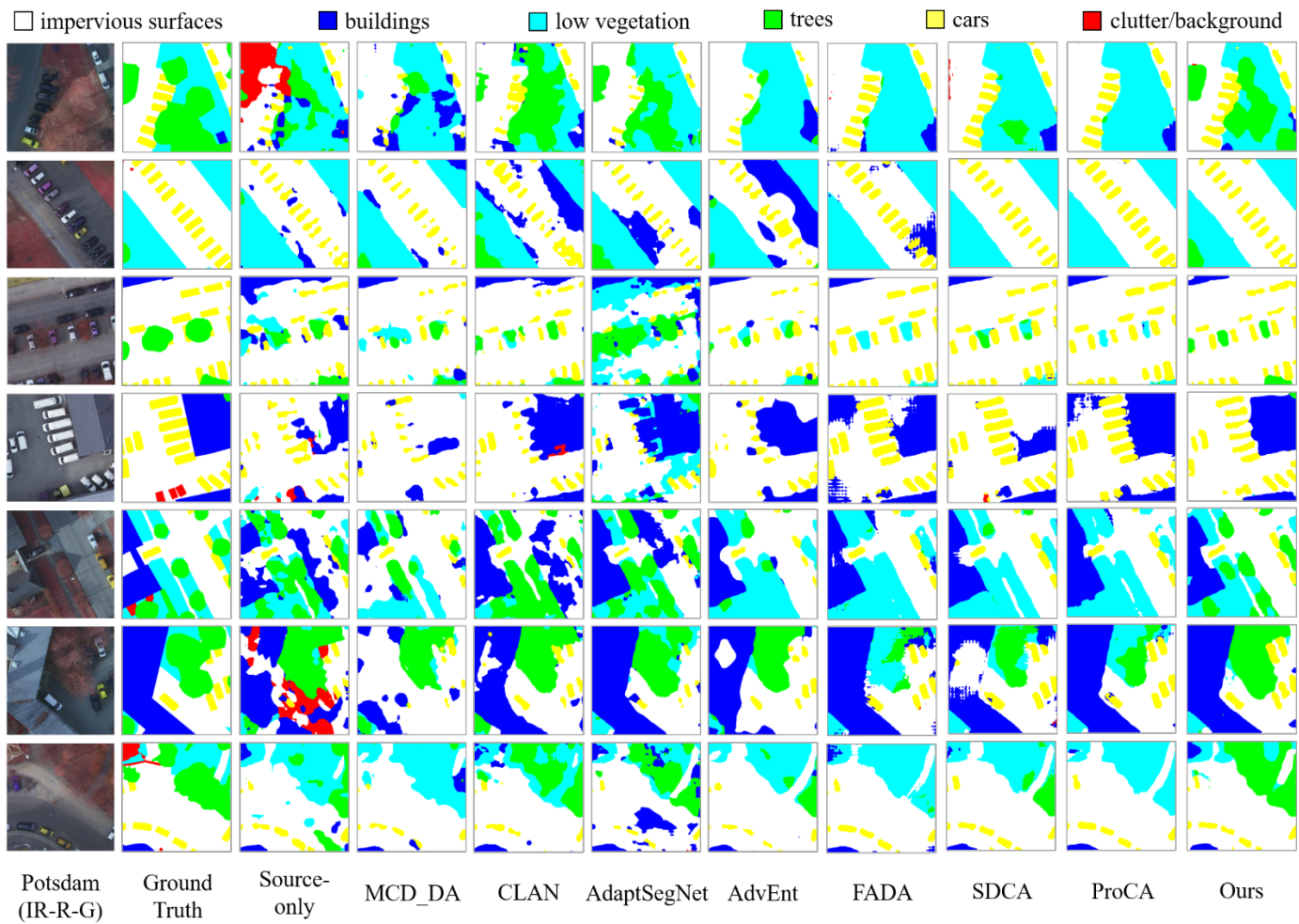

**Fig. 12.** Qualitative results on the V2P task.

### 2) Comparative Studies on V2P Task

We conduct the comparative experiment on the V2P task. The domain shift in this task is the same as the P2V_S task. The quantitative and qualitative evaluation results of the compared domain adaptation methods on this task are presented in Table III and Fig. 12.

From Table III, the source-only model obtains the worst performance with OA, MA, and mIoU values of 64.36%, 54.05%, and 41.22%, respectively. The performance of the state-of-the-art methods and our method is improved to a certain extent after adaptation. Our MemoryAdaptNet reaches the highest OA, MA, and mIoU values of 72.25%, 61.87%, and 49.82%, respectively, which obtains the improvement of 7.89%, 7.82%, and 8.60% on OA, MA, and mIoU compared with the source-only model, respectively. In Fig. 12, our MemoryAdaptNet achieves the best semantic segmentation performance with more accurate semantic information and smoother boundaries compared with other domain adaptation methods. These results demonstrate that our proposed MemoryAdaptNet can well deal with the V2P task.

*C. Ablation Experiment*

**The contribution of each component.** Our proposed MemoryAdaptNet benefits from the DFA and IDMA modules. We conduct four different ablation experiments (source-only, DFA, IDMA, DFA + IDMA) to evaluate the contribution of the MemoryAdaptNet's components to model performance. Source-only represents the segmentation network only trained on the source domain and is directly tested on the target domain. DFA represents the domain adaptation method that only uses the DFA branch to construct adversarial learning on the output space to align the feature between two domains. IDMA represents the model that without the D of DFA to implement adversarial learning on the output space. DFA + IDMA represents the model embedded with IDMA branch on the basis of DFA branch. The experiments are conducted on the P2V_S domain adaptation task, and the corresponding results are shown in Table IV.

As shown in Table IV, the DFA model employing the DFA branch yields an improvement of 3.32%, 4.72%, and 4.02% in OA, MA, and mIoU, respectively, compared with the source-only model. This finding verifies the effect of adversarial learning in the output space for UDA semantic segmentation. The IDMA model without the output space adversarial loss brings 7.69%, 5.95%, and 5.53% improvement in OA, MA, and mIoU, respectively, compared with the source-only model, even further surpassed the performance of the DFA model, which indicates that storing dataset-level feature information for source domain and target domain (even if the target feature is obtained without supervision) can significantly improve the semantic segmentation performance of target domain. Furthermore, when embedding the IDMA branch into the DFA branch (DFA +IDMA), the model yields the best performance with 77.87%, 77.05%, and 56.05% in OA, MA, and mIoU, respectively, which indicates that DFA and IDMA can promote each other to enhance the semantic segmentation performance of target domain. Notably, compared with the DFA model, the DFA + IDMA model yields an improvement of 7.36%, 11.78%, and 8.45% in OA, MA, and mIoU, which benefits from the invariant feature memory module in MemoryAdaptNet can integrate the pseudo invariant features of historical images to obtain category-level invariant domain-level features. Compared with the pseudo invariant features extracted from DFA, this feature can better represent the domain invariant features and maximize the transfer of knowledge learned in the source domain to the target domain.

**Memory Aggregation.** To verify the effectiveness of the memory aggregation module: CAMA used in our MemoryAdaptNet, we performed comparison experiments on CAMA and DCA used in [37] on P2V_S, P2V_D, and V2P tasks, the corresponding results are shown in Table V. Table V shows that our CAMA outperforms DCA on all three tasks. This is because the key features in DCA is obtained by matrix multiplication of $M$ and a category probability predicted by a classification head, since the predicted category probability may be misclassified, the key feature may introduce uncertain and incorrect information. However, our CAMA avoids the above problem by assigning $M$ as the key feature directly. Furthermore, our CAMA can adaptively enhance the presentation of current pseudo invariant features by seizing the prototype information beyond the current input images.

**Pseudo Label Filtering.** To verify the effectiveness of the entropy-based pseudo label filtering strategy used in our MemoryAdaptNet, we conduct comparative experiments with the existing pseudo label filtering schemes, the Probability-based Filtering represents the model with the highest score when probability threshold is 0.25, and the corresponding results are shown in Table VI. Compared to the existing pseudo label filtering schemes, our scheme achieves the best performance.

TABLE IV
ABLATION STUDY RESULTS (%) ON THE P2V_S TASK

|  | Methods | $F_1$ score | | | | | | OA | MA | mIoU |
|---|---|---|---|---|---|---|---|---|---|---|
|  |  | Imp. surf. | Building | Low veg. | Tree | Car | Clutter | | | |
|  | Source-only | 72.85 | 74.40 | 56.42 | 67.96 | 46.76 | 28.49 | 67.19 | 60.55 | 43.58 |
| Ablation | DFA | 74.77 | 78.29 | 61.18 | 64.98 | 53.66 | 45.06 | 70.51 | 65.27 | 47.60 |
| Study | IDMA | 79.02 | 85.84 | 60.87 | 78.19 | 52.28 | 19.47 | 74.88 | 66.50 | 49.11 |
|  | DFA + IDMA | **79.65** | **86.45** | **68.22** | **78.52** | **61.09** | **50.30** | **77.87** | **77.05** | **56.05** |

TABLE V
COMPARATIVE EXPERIMENTS WITH EXISTING MEMORY AGGREGATION SCHEMES.

| Task | Methods | $F_1$ score | | | | | | OA | MA | mIoU |
|---|---|---|---|---|---|---|---|---|---|---|
|  |  | Imp. surf. | Building | Low veg. | Tree | Car | Clutter | | | |
| P2V_S | Source-only | 72.85 | 74.40 | 56.42 | 67.96 | 46.76 | 28.49 | 67.19 | 60.55 | 43.58 |
|  | DCA | 78.75 | 85.03 | 62.91 | 77.09 | 59.60 | 43.53 | 75.90 | 74.52 | 52.96 |
|  | Ours | **79.65** | **86.45** | **68.22** | **78.52** | **61.09** | **50.30** | **77.87** | **77.05** | **56.05** |
| P2V_D | Source-only | 64.91 | 63.58 | 43.54 | 63.97 | 40.44 | 7.94 | 57.38 | 48.07 | 33.58 |
|  | DCA | **73.62** | 80.00 | 52.68 | 71.70 | 57.84 | 20.99 | 68.93 | 68.02 | 44.83 |
|  | Ours | 72.59 | **81.01** | **57.26** | 69.29 | **58.35** | **35.97** | **69.60** | **71.51** | **46.88** |
| V2P | Source-only | 71.32 | 69.45 | 60.97 | 54.67 | 69.10 | 5.47 | 64.36 | 54.05 | 41.22 |
|  | DCA | 78.55 | **80.71** | 65.95 | 66.02 | 70.93 | **15.61** | 71.93 | 61.18 | 48.75 |
|  | Ours | **79.70** | 77.84 | 67.58 | **64.06** | 76.54 | 16.15 | **72.25** | **61.87** | **49.82** |

TABLE VI
COMPARATIVE EXPERIMENTS WITH EXISTING PSEUDO LABEL FILTERING SCHEMES.

| Methods | $F_1$ score | | | | | | OA | MA | mIoU |
|---|---|---|---|---|---|---|---|---|---|
| | Imp. surf. | Building | Low veg. | Tree | Car | Clutter | | | |
| No Filtering | 78.20 | 86.07 | 65.99 | 77.42 | 59.41 | 49.57 | 76.81 | 76.70 | 54.56 |
| MUCSS[27] | 78.51 | 84.77 | 65.15 | 74.52 | 60.20 | **52.00** | 76.01 | 72.84 | 54.32 |
| ProCA [49] | 79.35 | 85.02 | 67.20 | 75.35 | 60.26 | 43.92 | 75.68 | 73.56 | 55.09 |
| Probability-based Filtering | 78.57 | 84.70 | 66.37 | 78.63 | 60.24 | 40.76 | 76.95 | **77.25** | 55.37 |
| Ours (Entropy-based) | **79.65** | **86.45** | **68.22** | **78.52** | **61.09** | 50.30 | **77.87** | 77.05 | **56.05** |

*D. Analysis on the Entropy Threshold of the Target Pseudo Label*

To investigate how the entropy threshold $\sigma$ of target pseudo label affect the proposed entropy-based pseudo label filtering strategy, we assess the performance of our proposed MemoryAdaptNet with different thresholds $\sigma$ on the P2V_S task, P2V_D task, and V2P task. The results of different thresholds $\sigma$ on the three tasks are shown in Tables VII, VIII, and IX.

For the threshold $\sigma$, $\sigma = 0$ represents that the pseudo label of target images does not participate in the update of invariant feature memory module; $\sigma = 1$ represents that the pseudo label of target images all are reserved for updating the invariant feature memory module without being filtered. When $0 < \sigma < 1$, the higher the $\sigma$ value, the more area of the pseudo label are reserved. $\sigma = 0.5$ represents that the pseudo label areas with entropy $\leq 0.5$ are reserved for the update of the invariant feature memory module, and the pseudo label areas with $0.5 < $ entropy $\leq 1$ are filtered. From Table VII, MemoryAdaptNet obtains the lowest score with OA of 75.07%, MA of 76.04%, and mIoU of 52.71% when the threshold $\sigma$ is 0. This finding is because when the pseudo label of target images does not participate in the update of invariant feature memory module, the invariant feature stored in the invariant feature memory module is more inclined to represent the source domain data distribution and cannot consider the target domain data distribution, resulting in poor segmentation performance of the model on the target domain. When the pseudo label of target domain participates in the update of invariant feature memory module ($\sigma > 0$), the segmentation performance of the model is improved. The simultaneous update of the pseudo invariant feature from the source and target domains enables the invariant feature stored in invariant feature memory module to represent the invariant data distribution of source and target domains. Similar experimental phenomena appear in the P2V_D task and V2P task, as shown in Tables VIII and IX. When the threshold $\sigma > 0$, MemoryAdaptNet obtains the highest score with OA of 77.87%, MA of 77.05%, and mIoU of 56.05% when the threshold $\sigma$ is 0.5 on the P2V_S task. For the other tasks, MemoryAdaptNet obtains the highest score with OA of 69.60%, MA of 71.51%, and mIoU of 46.88% when the threshold $\sigma$ is 0.25 on the P2V_D task and obtains the highest score with OA of 72.25%, MA of 61.87%, and mIoU of 49.82% when the threshold $\sigma$ is 0.5 on the V2P task.

*E. Effects of Different Data Augmentation Strategies*

In this section, we discuss the effects of different data augmentation strategies on the UDA semantic segmentation of HRS imagery. We conducted three experiments with different data augmentation strategies (w/o aug., affine aug., color-space aug.) on the domain adaptation task from P2V_S and the domain adaptation task from P2V_D. The corresponding results can be found in Table X. In Table X, the w/o aug. represents that no data augmentation strategy is utilized during the MemoryAdaptNet training. The affine aug. represents that the data augmentation strategy of affine transformation, such as horizontal flip, vertical flip, random rotate, and shift scale rotate, is used during the MemoryAdaptNet training. The color-space aug. represents the data augmentation strategy based on the color-space transformation, such as Gaussian noise, Gaussian blur, random brightness, and random contrast, is used during the MemoryAdaptNet training.

As shown in Table X, the w/o aug. model achieves the performance of 74.04%, 68.78%, and 50.47% in OA, MA, and mIoU on the P2V_S task and the performance of 61.76%, 61.98%, and 39.78% in OA, MA, and mIoU on the P2V_D task, respectively. Compared with the w/o aug. model, the performance of the color-space aug. model decreases to varying degrees on the two domain adaptation tasks. Specifically, the OA, MA, and mIoU on the P2V_S task decreased by 3.39%, 0.99%, and 3.17%, respectively, and the OA, MA, and mIoU on the P2V_D task decreased by 2.48%, 12.61%, and 4.18%, respectively, which indicates that the data augmentation strategy based on color-space transformation has a negative effect on the domain adaptation semantic segmentation of HRS image. This finding may be because the UDA semantic segmentation task intends to narrow the domain gap between the source domain and the target domain. However, the data augmentation method based on color-space transformation increases the discrepancy in appearance and data distribution between the two domains by adding Gaussian noise, Gaussian blur or random brightness to the source domain and target domain images randomly. Thus, the performance of the color-space aug. model is lower than that of the w/o aug. model. The negative effect of the data augmentation method based on color-space transformation on the P2V_D task is greater than that on the P2V_S task. This finding is because the distribution difference caused by color-space transformation on datasets with large color-space differences (existing in P2V_D task) is greater than that on datasets with small color-space difference (existing in P2V_S task). Compared with the w/o aug. model, the affine aug. model makes performance improvement on the two domain adaptation tasks. Specifically, this model achieves 3.83%, 8.27%, and 5.58% improvement of OA, MA, and mIoU on the P2V_S task, and 7.84%, 9.53%, and 7.10% improvement of OA, MA, and mIoU on the P2V_D task, respectively. This result shows that appropriate affine transformation during model training can promote the performance of domain adaptation semantic segmentation of HRS imagery.

TABLE VII
PARAMETER ANALYSIS OF THE TARGET PSEUDO LABEL ENTROPY THRESHOLD ON THE P2V_S TASK

| σ | | | $F_1$ score | | | | OA | MA | mIoU |
|---|---|---|---|---|---|---|---|---|---|
| | Imp. surf. | Building | Low veg. | Tree | Car | Clutter | | | |
| 0.00 | 77.76 | 84.75 | 62.42 | 77.23 | 60.90 | 36.67 | 75.07 | 76.04 | 52.71 |
| 0.25 | 77.46 | 86.44 | 65.77 | 77.34 | 59.66 | **51.86** | 76.46 | **78.42** | 54.82 |
| 0.50 | **79.65** | **86.45** | **68.22** | **78.52** | **61.09** | 50.30 | **77.87** | 77.05 | **56.05** |
| 0.75 | 78.39 | 85.65 | 65.33 | 78.24 | 58.79 | 40.74 | 76.88 | 76.28 | 53.22 |
| 1.00 | 78.20 | 86.07 | 65.99 | 77.42 | 59.41 | 49.57 | 76.81 | 76.70 | 54.56 |

TABLE VIII
PARAMETER ANALYSIS OF THE TARGET PSEUDO LABEL ENTROPY THRESHOLD ON THE P2V_D TASK

| σ | | | $F_1$ score | | | | OA | MA | mIoU |
|---|---|---|---|---|---|---|---|---|---|
| | Imp. surf. | Building | Low veg. | Tree | Car | Clutter | | | |
| 0.00 | **74.96** | 78.55 | 51.59 | 67.25 | 57.51 | 19.88 | 66.82 | 67.74 | 43.57 |
| 0.25 | 72.59 | 81.01 | 57.26 | 69.29 | **58.35** | **35.97** | **69.60** | **71.51** | **46.88** |
| 0.50 | 72.80 | 79.74 | 47.84 | **70.38** | 58.12 | 26.52 | 67.77 | 67.39 | 44.25 |
| 0.75 | 74.49 | 81.14 | 54.47 | 62.82 | 58.33 | 29.32 | 66.91 | 71.23 | 44.92 |
| 1.00 | 73.79 | **82.15** | 54.48 | 69.44 | 56.37 | 23.31 | 68.08 | 68.76 | 45.21 |

TABLE IX
PARAMETER ANALYSIS OF THE TARGET PSEUDO LABEL ENTROPY THRESHOLD ON THE V2P TASK

| σ | | | $F_1$ score | | | | OA | MA | mIoU |
|---|---|---|---|---|---|---|---|---|---|
| | Imp. surf. | Building | Low veg. | Tree | Car | Clutter | | | |
| 0.00 | 76.96 | 77.19 | 64.78 | 56.35 | 75.87 | 8.97 | 69.41 | 59.38 | 46.39 |
| 0.25 | 80.58 | 83.55 | 64.46 | 65.80 | 62.01 | 13.82 | 72.94 | 59.70 | 48.03 |
| 0.50 | **79.70** | **77.84** | 67.58 | **64.06** | **76.54** | **16.15** | 72.25 | **61.87** | **49.82** |
| 0.75 | 79.13 | 77.71 | 67.34 | 63.70 | 75.96 | 7.09 | **72.39** | 61.08 | 49.19 |
| 1.00 | 78.60 | 78.94 | 67.10 | 63.55 | 75.02 | 6.57 | 72.01 | 60.89 | 48.40 |

TABLE X
QUANTITATIVE RESULTS (%) OF DIFFERENT DATA AUGMENTATION STRATEGIES ON THE P2V_S TASK AND P2V_D TASK

| Methods | | | | $F_1$ score | | | | OA | MA | mIoU |
|---|---|---|---|---|---|---|---|---|---|---|
| | | Imp. surf. | Building | Low veg. | Tree | Car | Clutter | | | |
| P2V_S task | w/o aug. | 76.67 | 80.04 | 64.21 | 64.44 | 48.04 | 47.74 | 74.04 | 68.78 | 50.47 |
| | color-space aug. | 75.68 | 80.19 | 58.70 | 60.48 | 52.65 | 36.86 | 70.65 | 67.79 | 47.30 |
| | affine aug. | **79.65** | **86.45** | **68.22** | **78.52** | **61.09** | 50.30 | **77.87** | **77.05** | **56.05** |
| P2V_D task | w/o aug. | 68.51 | 76.13 | 51.07 | 46.83 | 51.87 | **46.95** | 61.76 | 61.98 | 39.78 |
| | color-space aug. | 69.49 | 76.29 | 48.53 | 49.34 | 47.66 | 5.05 | 59.28 | 49.37 | 35.60 |
| | affine aug. | **72.59** | **81.01** | **57.26** | 69.29 | **58.35** | 35.97 | **69.60** | **71.51** | **46.88** |

## V. CONCLUSION AND FUTURE WORK

To address variation of data distribution from different domain in semantic segmentation of HRS imagery, we propose a novel UDA semantic segmentation network named MemoryAdaptNet. Essentially, the MemoryAdaptNet use an invariant feature memory module to refrain from the negative tendency of only representing the variant feature of current limited inputs in an UDA with adversarial learning scheme. Given that the memory module can capture prototype information, the MemoryAdaptNet is capable of storing the invariant domain-level prototype information. For further augmenting the feature representation of input image from target domain, we aggregate the invariant domain-level memory to the current pseudo invariant feature produced by the feature extractor with a category attention-driven module. Thus, the proposed model can consequently enhance the performance of semantic segmentation of HRS imagery in the target domain. The results of extensive experiments indicate the effectiveness of our MemoryAdaptNet, which favorably outperforms the baseline model and the state-of-the-art models.

In the future, we aim to further improve the MemoryAdaptNet from the following aspects: 1) performing adversarial learning at the output space with different feature levels by considering that different level features of DCNN excavate different semantic and other detailed information, and 2) exploring a pseudo label filtering strategy based on adaptive entropy threshold rather than manually setting the entropy threshold for reducing the manual intervention and workload.

## REFERENCES

[1] Cheng, G., J. Han, and X. Lu, *Remote sensing image scene classification: Benchmark and state of the art.* Proceedings of the IEEE, 2017. **105**(10): p. 1865-1883.

[2] Hu, F., et al., *Transferring deep convolutional neural networks for the scene classification of high-resolution remote sensing imagery.* Remote Sensing, 2015. **7**(11): p. 14680-14707.

[3] Chen, J., et al., *Geospatial relation captioning for high-spatial-resolution images by using an attention-based neural network.* International Journal of Remote Sensing, 2019. **40**(16): p. 6482-6498.

[4] Zhao, R., Z. Shi, and Z. Zou, *High-resolution remote sensing image captioning based on structured attention.* IEEE Transactions on Geoscience and Remote Sensing, 2021. **60**: p. 1-14.

[5] Chen, J., et al., *Multi-scale spatial and channel-wise attention for improving object detection in remote sensing imagery*. IEEE Geoscience and Remote Sensing Letters, 2019. **17**(4): p. 681-685.

[6] Li, K., et al., *Object detection in optical remote sensing images: A survey and a new benchmark*. ISPRS Journal of Photogrammetry and Remote Sensing, 2020. **159**: p. 296-307.

[7] Chen, J., et al., *SMAF-Net: Sharing multiscale adversarial feature for high-resolution remote sensing imagery semantic segmentation*. IEEE Geoscience and Remote Sensing Letters, 2020. **18**(11): p. 1921-1925.

[8] Zhang, J., et al., *Multi-scale context aggregation for semantic segmentation of remote sensing images*. Remote Sensing, 2020. **12**(4): p. 701.

[9] Long, J., E. Shelhamer, and T. Darrell. *Fully convolutional networks for semantic segmentation*. in *Proceedings of the IEEE conference on computer vision and pattern recognition*. 2015. p. 3431-3440.

[10] Zuo, T., J. Feng, and X. Chen. *HF-FCN: Hierarchically fused fully convolutional network for robust building extraction*. in *Asian Conference on Computer Vision*. 2016. Springer. p. 291-302.

[11] Yang, Y., et al., *M-FCN: Effective fully convolutional network-based airplane detection framework*. IEEE Geoscience and Remote Sensing Letters, 2017. **14**(8): p. 1293-1297.

[12] Badrinarayanan, V., A. Kendall, and R. Cipolla, *Segnet: A deep convolutional encoder-decoder architecture for image segmentation*. IEEE transactions on pattern analysis and machine intelligence, 2017. **39**(12): p. 2481-2495.

[13] Ronneberger, O., P. Fischer, and T. Brox. *U-net: Convolutional networks for biomedical image segmentation*. in *International Conference on Medical image computing and computer-assisted intervention*. 2015. Springer. p. 234-241.

[14] Zhao, H., et al. *Pyramid scene parsing network*. in *Proceedings of the IEEE conference on computer vision and pattern recognition*. 2017. p. 2881-2890.

[15] Chen, L.-C., et al., *Deeplab: Semantic image segmentation with deep convolutional nets, atrous convolution, and fully connected crfs*. IEEE transactions on pattern analysis and machine intelligence, 2017. **40**(4): p. 834-848.

[16] Goodfellow, I., et al., *Generative adversarial nets*. Advances in neural information processing systems, 2014. **27**.

[17] Hoffman, J., et al. *Cycada: Cycle-consistent adversarial domain adaptation*. in *International conference on machine learning*. 2018. PMLR. p. 1989-1998.

[18] Wu, Z., et al. *Dcan: Dual channel-wise alignment networks for unsupervised scene adaptation*. in *Proceedings of the European Conference on Computer Vision (ECCV)*. 2018. p. 518-534.

[19] Li, Y., L. Yuan, and N. Vasconcelos. *Bidirectional learning for domain adaptation of semantic segmentation*. in *Proceedings of the IEEE/CVF Conference on Computer Vision and Pattern Recognition*. 2019. p. 6936-6945.

[20] Hoffman, J., et al., *Fcns in the wild: Pixel-level adversarial and constraint-based adaptation*. arXiv preprint arXiv:1612.02649, 2016.

[21] Chen, Y.-H., et al. *No more discrimination: Cross city adaptation of road scene segmenters*. in *Proceedings of the IEEE International Conference on Computer Vision*. 2017. p. 1992-2001.

[22] Tsai, Y.-H., et al. *Learning to adapt structured output space for semantic segmentation*. in *Proceedings of the IEEE conference on computer vision and pattern recognition*. 2018. p. 7472-7481.

[23] Luo, Y., et al. *Taking a closer look at domain shift: Category-level adversaries for semantics consistent domain adaptation*. in *Proceedings of the IEEE/CVF Conference on Computer Vision and Pattern Recognition*. 2019. p. 2507-2516.

[24] Vu, T.-H., et al. *Advent: Adversarial entropy minimization for domain adaptation in semantic segmentation*. in *Proceedings of the IEEE/CVF Conference on Computer Vision and Pattern Recognition*. 2019. p. 2517-2526.

[25] Wittich, D. and F. Rottensteiner, *Appearance based deep domain adaptation for the classification of aerial images*. ISPRS Journal of Photogrammetry and Remote Sensing, 2021. **180**: p. 82-102.

[26] Zhang, L., et al., *Stagewise unsupervised domain adaptation with adversarial self-training for road segmentation of remote-sensing images*. IEEE Transactions on Geoscience and Remote Sensing, 2021. **60**: p. 1-13.

[27] Li, Y., et al., *Learning deep semantic segmentation network under multiple weakly-supervised constraints for cross-domain remote sensing image semantic segmentation*. ISPRS Journal of Photogrammetry and Remote Sensing, 2021. **175**: p. 20-33.

[28] Chen, J., et al., *Unsupervised Domain Adaptation for Semantic Segmentation of High-Resolution Remote Sensing Imagery Driven by Category-Certainty Attention*. IEEE Transactions on Geoscience and Remote Sensing, 2022. **60**: p. 1-15.

[29] Geng, B., D. Tao, and C. Xu, *DAML: Domain adaptation metric learning*. IEEE Transactions on Image Processing, 2011. **20**(10): p. 2980-2989.

[30] Zellinger, W., et al., *Central moment discrepancy (cmd) for domain-invariant representation learning*. arXiv preprint arXiv:1702.08811, 2017.

[31] Elshamli, A., et al., *Domain adaptation using representation learning for the classification of remote sensing images*. IEEE Journal of Selected Topics in Applied Earth Observations and Remote Sensing, 2017. **10**(9): p. 4198-4209.

[32] Song, S., et al., *Domain adaptation for convolutional neural networks-based remote sensing scene classification*. IEEE Geoscience and Remote Sensing Letters, 2019. **16**(8): p. 1324-1328.

[33] Zhang, J., et al., *Domain adaptation based on correlation subspace dynamic distribution alignment for remote sensing image scene classification*. IEEE Transactions on Geoscience and Remote Sensing, 2020. **58**(11): p. 7920-7930.

[34] Chunseong Park, C., B. Kim, and G. Kim. *Attend to you: Personalized image captioning with context sequence memory networks*. in *Proceedings of the IEEE conference on computer vision and pattern recognition*. 2017. p. 895-903.

[35] Chen, Y., et al. *Memory enhanced global-local aggregation for video object detection*. in *Proceedings of the IEEE/CVF Conference on Computer Vision and Pattern Recognition*. 2020. p. 10337-10346.

[36] Oh, S.W., et al. *Video object segmentation using space-time memory networks*. in *Proceedings of the IEEE/CVF International Conference on Computer Vision*. 2019. p. 9226-9235.

[37] Jin, Z., et al. *Mining contextual information beyond image for semantic segmentation*. in *Proceedings of the IEEE/CVF International Conference on Computer Vision*. 2021. p. 7231-7241.

[38] Alonso, I., et al. *Semi-supervised semantic segmentation with pixel-level contrastive learning from a class-wise memory bank*. in *Proceedings of the IEEE/CVF International Conference on Computer Vision*. 2021. p. 8219-8228.

[39] He, K., et al. *Deep residual learning for image recognition*. in *Proceedings of the IEEE conference on computer vision and pattern recognition*. 2016. p. 770-778.

[40] Deng, J., et al. *Imagenet: A large-scale hierarchical image database*. in *2009 IEEE conference on computer vision and pattern recognition*. 2009. Ieee. p. 248-255.

[41] Wu, Z., et al. *Unsupervised feature learning via non-parametric instance discrimination*. in *Proceedings of the IEEE conference on computer vision and pattern recognition*. 2018. p. 3733-3742.

[42] He, K., et al. *Momentum contrast for unsupervised visual representation learning*. in *Proceedings of the IEEE/CVF conference on computer vision and pattern recognition*. 2020. p. 9729-9738.

[43] Rottensteiner, F., et al., *The ISPRS benchmark on urban object classification and 3D building reconstruction*. ISPRS Annals of the Photogrammetry, Remote Sensing and Spatial Information Sciences I-3 (2012), Nr. 1, 2012. **1**(1): p. 293-298.

[44] Bottou, L., *Large-scale machine learning with stochastic gradient descent*, in *Proceedings of COMPSTAT'2010*. 2010, Springer. p. 177-186.

[45] Kingma, D.P. and J. Ba, *Adam: A method for stochastic optimization*. arXiv preprint arXiv:1412.6980, 2014.

[46] Saito, K., et al. *Maximum classifier discrepancy for unsupervised domain adaptation*. in *Proceedings of the IEEE conference on computer vision and pattern recognition*. 2018. p. 3723-3732.

[47] Wang, H., et al. *Classes matter: A fine-grained adversarial approach to cross-domain semantic segmentation*. in *European conference on computer vision*. 2020. Springer. p. 642-659.

[48] Li, S., et al., *Semantic distribution-aware contrastive adaptation for semantic segmentation*. arXiv preprint arXiv:2105.05013, 2021.

[49] Zhang B, Chen T, Wang B. *Curriculum-style local-to-global adaptation for cross-domain remote sensing image segmentation*[J]. IEEE Transactions on Geoscience and Remote Sensing, 2021, 60: 1-12.

[50] Jiang, Z., et al., *Prototypical Contrast Adaptation for Domain Adaptive Semantic Segmentation*. arXiv preprint arXiv:2207.06654, 2022.